%% file: main.tex
\documentclass[10pt,twocolumn,letterpaper]{article}

\usepackage[pagenumbers]{cvpr} % To force page numbers, e.g. for an arXiv version

\input{preamble}

\definecolor{cvprblue}{rgb}{0.21,0.49,0.74}
\usepackage[pagebackref,breaklinks,colorlinks,citecolor=cvprblue]{hyperref}

\def\paperID{*****} % *** Enter the Paper ID here
\def\confName{CVPR}
\def\confYear{2024}

\usepackage{subcaption}
\usepackage{booktabs}
\usepackage{multirow}
\usepackage{newfloat}
\usepackage{comment}

\title{Automated Virtual Product Placement and \\ 
Assessment in Images using Diffusion Models}

\author{Mohammad Mahmudul Alam\thanks{The author performed this work as an intern at Amazon Web Services (AWS). Accepted at 6\textsuperscript{th} AI for Content Creation (AI4CC) workshop at CVPR 2024. (Preprint)}\\
University of Maryland, Baltimore County\\
Baltimore, MD, USA\\
{\tt\small m256@umbc.edu}
\and
Negin Sokhandan\\
Amazon Web Services (AWS)\\
Santa Clara, CA, USA\\
{\tt\small ngnsl@amazon.com}
\and
Emmett Goodman\\
Amazon Web Services (AWS)\\
Santa Clara, CA, USA\\
{\tt\small edmgood@amazon.com}
}

\begin{document}

\maketitle

\begin{abstract}
In Virtual Product Placement (VPP) applications, the discrete integration of specific brand products into images or videos has emerged as a challenging yet important task. This paper introduces a novel three-stage fully automated VPP system. In the first stage, a language-guided image segmentation model identifies optimal regions within images for product inpainting. In the second stage, Stable Diffusion (SD), fine-tuned with a few example product images, is used to inpaint the product into the previously identified candidate regions. The final stage introduces an `Alignment Module', which is designed to effectively sieve out low-quality images. Comprehensive experiments demonstrate that the Alignment Module ensures the presence of the intended product in every generated image and enhances the average quality of images by $35\%$. The results presented in this paper demonstrate the effectiveness of the proposed VPP system, which holds significant potential for transforming the landscape of virtual advertising and marketing strategies.
\end{abstract}

\section{Introduction}
Virtual Product Placement (VPP) refers to the unobtrusive, digital integration of branded products into visual content, which is often employed as a stealth marketing strategy \cite{mcdonnell2010virtual}. Advertising solutions utilizing VPP have significant appeal due to their high customizability, effectiveness across diverse customer bases, and quantifiable efficiency. Previous research underscores the impact of product placement within realms such as virtual reality \cite{wang2019influence} and video games \cite{glass2007effectiveness}. With the recent advancements in generative AI technologies, the potential for product placement has been further expanded through the utilization of diffusion models. Significant research has focused on the development of controlled inpainting via diffusion models, albeit largely without an explicit emphasis on advertising applications \cite{avrahami2022blended, kawar2023imagic, li2023gligen}. However, these methods can be fine-tuned with a small set of 4 to 5 product sample images to generate high-quality advertising visual content.

\begin{figure}[!t]
\centering
\begin{subfigure}{0.28\linewidth}
    \includegraphics[width=\linewidth]{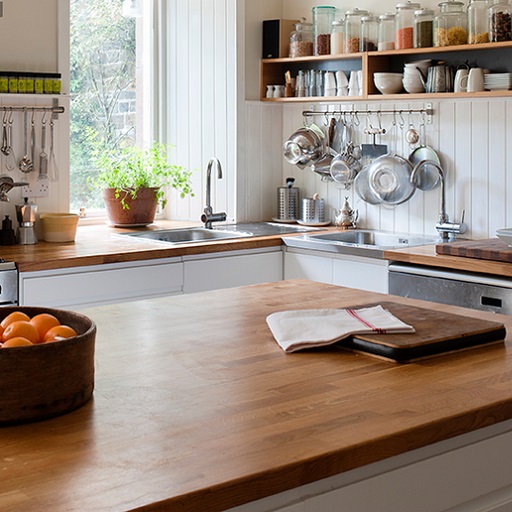}
    \caption{Background}
\end{subfigure}
\hspace{3mm}
\begin{subfigure}{0.28\linewidth}
    \includegraphics[width=\linewidth]{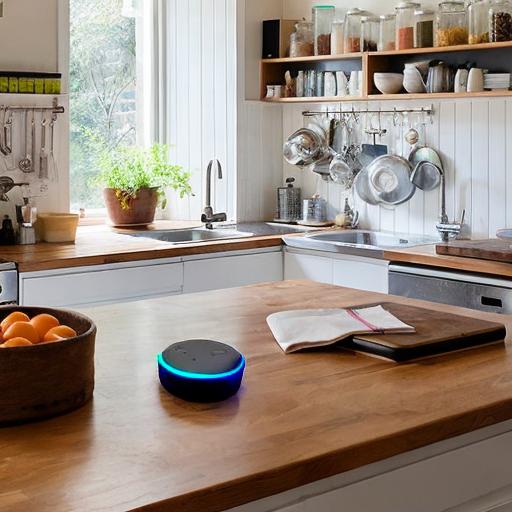}
    \caption{Inpainting}
\end{subfigure}
\caption{An illustration of the proposed VPP system with an Amazon Echo Dot device. The input background image is shown in (a), and the inpainted output image is shown in (b) where an Amazon Echo Dot device is placed on the kitchen countertop by automatic identification of optimal location.}
\label{fig:demo}
\end{figure}

In this paper, we propose a novel, three-stage, fully automated system that carries out semantic inpainting of products by fine-tuning a pre-trained Stable Diffusion (SD) model \cite{sd}. In the first stage, a suitable location is identified for product placement using visual question answering and text-conditioned instant segmentation. The output of this stage is a binary mask highlighting the identified location. Subsequently, this masked region undergoes inpainting using a fine-tuned SD model. This SD model is fine-tuned by DreamBooth \cite{dreambooth} approach utilizing a few sample images of the product along with a unique identifier text prompt. Finally, the quality of the inpainted image is evaluated by a proposed Alignment Module, a discriminative method that measures the image quality, or the alignment of the generated image with human expectations. An illustration of the proposed VPP system is presented in \autoref{fig:demo} with an Amazon Echo Dot device.
\par 
Controlled inpainting of a specific product is a challenging task. For example, the model may fail to inpaint the intended object at all. If a product is indeed introduced through inpainting, the product created may not be realistic and may display distortions of shape, size, or color. Similarly, the background surrounding the inpainted product may be altered in such a way that it either meaningfully obscures key background elements or even completely changes the background image. This becomes especially problematic when the background images contain human elements, as models can transform them into disturbing visuals. As a result, the proposed Alignment Module is designed to address these complications, with its primary focus being on the appearance, quality, and size of the generated product.
\par 
To exert control over the size of the generated product, morphological transformations, specifically erosion, and dilation, are employed. By adjusting the size of the mask through dilation or erosion, the size of the inpainted product can be effectively increased or decreased. This allows the system to generate a product of an appropriate size. 
\par 
In summary, the main contributions of this paper are twofold. The first pertains to the design of a fully automated Virtual Product Placement (VPP) system capable of generating high-resolution, customer-quality visual content. The second involves the development of a discriminative method that automatically eliminates subpar images, premised on the content, quality, and size of the product generated.
\par 
The remainder of this paper is organized as follows. In \autoref{sec:related} we will delve into the related literature, with a specific emphasis on semantic inpainting methods utilizing diffusion models, and \autoref{sec:contribute} will highlight the broad contributions of the paper. Next, the proposed end-to-end pipeline for automatic VPP will be discussed in \autoref{sec:method}. This includes a detailed examination of the three primary stages of the solution, along with the three sub-modules of the Alignment Module. Thereafter, we will elucidate the experimental design and evaluation methodologies adopted and report the corresponding results in \autoref{sec:experiment}. Subsequently, deployment strategy and web application design will be explained in \autoref{sec:production}. Finally, the paper will conclude with an outline of the identified limitations of our proposed methodology in \autoref{sec:conclusion}, complemented by a discussion on potential avenues for future research.

\section{Related Works} \label{sec:related}
Recently, there has been significant progress in developing semantic or localized image editing using diffusion models - largely without an explicit focus on digital marketing. Nevertheless, new generative AI approaches promise significant advances in VPP technology. For instance, in Blended Diffusion \cite{avrahami2022blended}, the authors proposed a method of localized image editing using image masking and natural language. The area of interest is first masked and then modified using a text prompt. The authors employed a pre-trained CLIP model \cite{clip} along with pre-trained Denoising Diffusion Probabilistic Models (DDPM) \cite{ddpm} to generate natural images in the area of interest. 
\par 
Similar to Blended Diffusion, Couairon et. al. \cite{couairon2022diffedit} proposed a method of semantic editing with a mask using a diffusion model. However, instead of taking the mask from the user, the mask is generated automatically. Nevertheless, a text query input from the user is utilized to generate the mask. The difference in noise estimates, as determined by the diffusion model based on the reference text and the query text, is calculated. This difference is then used to infer the mask. The image is noised iteratively during the forward process and in the reverse Denoising Diffusion Implicit Model (DDIM) \cite{ddim} steps, the denoised image is interpolated with the same step output of the forward process using masking.
\par
Paint by Word proposed by Bau et. al. \cite{bau2021paint}, is also similar, however instead of a diffusion model they utilized a Generative Adversarial Networks (GAN) \cite{creswell2018generative} with a mask for semantic editing guided by text. On the other hand, Imagic \cite{kawar2023imagic} also performs text-based semantic editing on images using a diffusion model but without using any mask. Their approach consists of three steps. In the beginning, text embedding for a given image is optimized. Then the generative diffusion model is optimized for the given image with fixed-optimized text embedding. Finally, the target and optimized embedding are linearly interpolated to achieve input image and target text alignment. 
Likewise, a semantic editing method using a pre-trained text-conditioned diffusion model focusing on the mixing of two concepts is proposed by \cite{liew2022magicmix}. In this method, a given image is noised for several steps and then denoised with text condition. During the denoising process, the output of a denoising stage is also linearly interpolated with the output of a forward noise mixing stage. 
\par 
Hertz et. al. \cite{hertz2022prompt} took a different approach to semantic image editing where text and image embeddings are fused using cross-attention. The cross-attention maps are incorporated with the Imagen diffusion model \cite{saharia2022photorealistic}. However, instead of editing any given image, their approach edits a generated image using a text prompt which lacks any interest when VPP is concerned. Alternatively, Stochastic Differential Edit (SDEdit) \cite{meng2021sdedit} synthesizes images from stroke paintings and can edit images based on stroke images. For image synthesis, coarse colored strokes are used and for editing, colored stroke on real images or image patches on target images is used as a guide. It adds Gaussian noise to an image guide of a specific standard deviation and then solves the corresponding Stochastic Differential Equations (SDE) to produce the synthetic or edited image. 
\par 
To generate images from a prompt in a controlled fashion and to gain more control over the generated image, Li et. al proposed grounded text-to-image generation (GLIGEN) \cite{li2023gligen}. It feeds the model the embedding of the guiding elements such as bounding boxes, key points, or semantic maps. Using the same guiding components, inpainting can be performed in a target image. 
\par
DreamBooth \cite{dreambooth} fine-tunes the pre-trained diffusion model to expand the dictionary of the model for a specific subject. Given a few examples of the subject, a diffusion model such as Imagen \cite{saharia2022photorealistic} is fine-tuned using random samples generated by the model itself and new subject images by optimizing a reconstruction loss. The new subject images are conditioned using a text prompt with a unique identifier. Fine-tuning a pre-trained diffusion model with a new subject is of great importance in the context of VPP. Therefore, in this paper DreamBooth approach is utilized to expand the model's dictionary by learning from a few sample images of the product.

\section{Contributions} \label{sec:contribute}
In this paper, a method of automated virtual product placement and assessment in images using diffusion models is designed. Our broad contributions are as follows:
\begin{enumerate}
    \item We introduce a novel fully automated VPP system that carries out automatic semantic inpainting of the product in the optimal location using language-guided segmentation and fine-tuned stable diffusion models.
    
    \item We proposed a cascaded three-stage assessment module named `Alignment Module' designed to sieve out low-quality images that ensure the presence of the intended product in every generated output image.
    
    \item Morphological transformations are employed such as dilation and erosion to adjust the size of the mask, therefore, to increase or decrease the size of the inpainted product allowing generating a product of appropriate size. 
    
    \item Experiments are performed to validate the results by blind evaluation of the generated images with and without the Alignment module resulting in $35\%$ improvement in average quality.
    
    \item The inpainted product generated by the proposed system is not only qualitatively more realistic compared to the previous inpainting approach \cite{yang2023paint} but also shows a superior quantitative CLIP score.
\end{enumerate}

\section{Methodology} \label{sec:method}

\begin{figure}[!htbp]
\centering
\includegraphics[width=\columnwidth]{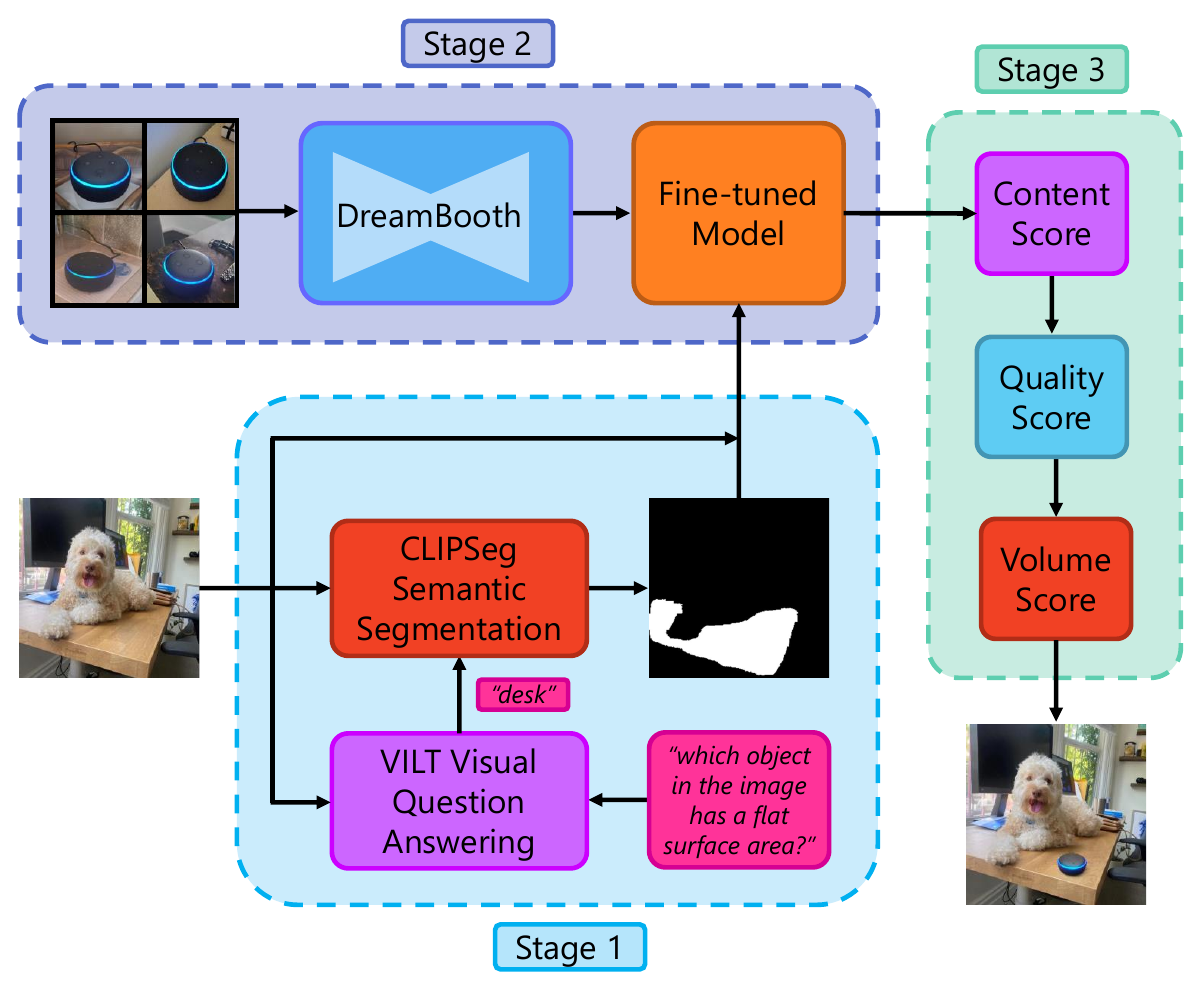}
\caption{The block diagram of the proposed solution for the VPP system where each of the three stages is distinguished by varied color blocks. In stage 1, a suitable placement for product inpainting is determined by creating a mask using CLIPSeg and VILT models. Next, in stage 2, semantic inpainting is performed in the masked area using the fine-tuned DreamBooth model. Finally, stage 3 contains the cascaded sub-modules of the Alignment Module to discard low-quality images.}
\label{fig:block_diagram}
\end{figure}

\subsection{Proposed Method} 
For semantic inpainting, we utilized the DreamBooth algorithm \cite{dreambooth} to fine-tune stable diffusion using five representative images of the product and a text prompt with a unique identifier. Even with a limited set of five sample images, the fine-tuned DreamBooth model was capable of generating images of the product integrated with its background. 
\par 
Nevertheless, when inpainting was conducted with this fine-tuned model, the resulting quality of the inpainted product was significantly compromised. To enhance the quality of the product in the inpainted image, we augmented the sample images through random scaling and random cropping, consequently generating a total of 1,000 product images used to fine-tune SD.

\subsection{Product Localization Module}
The proposed VPP system operates in three stages. A core challenge in product placement lies in pinpointing a suitable location for the item within the background. In the first stage, this placement is indicated via the generation of a binary mask. To automate this masking process, we leveraged the capabilities of the Vision and Language Transformer (ViLT) Visual Question Answering (VQA) model \cite{vilt} in conjunction with the Contrastive Language-Image Pretraining (CLIP) \cite{clip}-based semantic segmentation method, named CLIPSeg \cite{clipseg}. Notably, each product tends to have a prototypical location for its placement. For example, an optimal location for an Amazon Echo Dot device is atop a flat surface, such as a desk or table. Thus, by posing a straightforward query to the VQA model, such as "Which object in the image has a flat surface area?", we can pinpoint an appropriate location for the product. Subsequently, the identified location's name is provided to the CLIPSeg model, along with the input image, resulting in the generation of a binary mask for the object.

\subsection{Product Inpainting Module}
In the second stage, the input image and the generated binary mask are fed to the fine-tuned DreamBooth model to perform inpainting on the masked region. Product inpainting presents several challenges: the product might not manifest in the inpainted region; if it does, its quality could be compromised or distorted, and its size might be disproportionate to the surrounding context. To systematically detect these issues, we introduce the third stage: the Alignment Module.

\begin{figure}[!t]
\centering
\begin{subfigure}[!htbp]{0.94\columnwidth}
    \centering
    \includegraphics[width=\columnwidth]{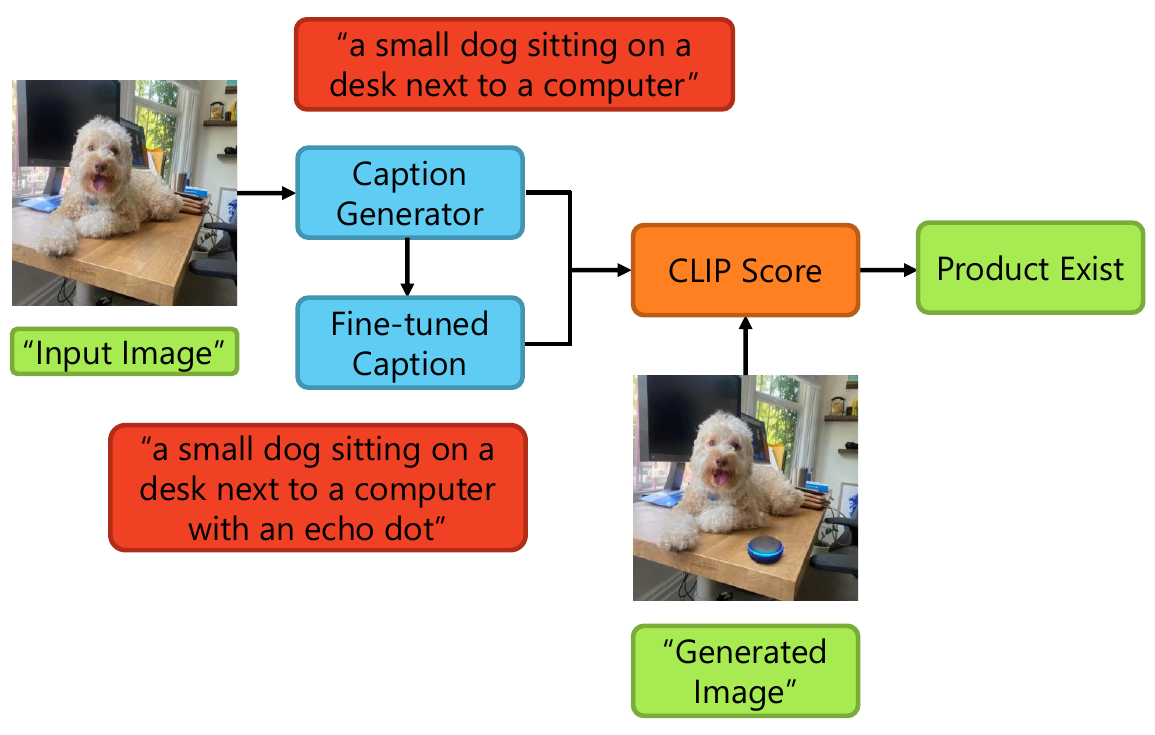}
    \caption{Content Sub-module}
    \vspace{5pt}
\end{subfigure}

\begin{subfigure}[!htbp]{0.94\columnwidth}
    \centering
    \includegraphics[width=\columnwidth]{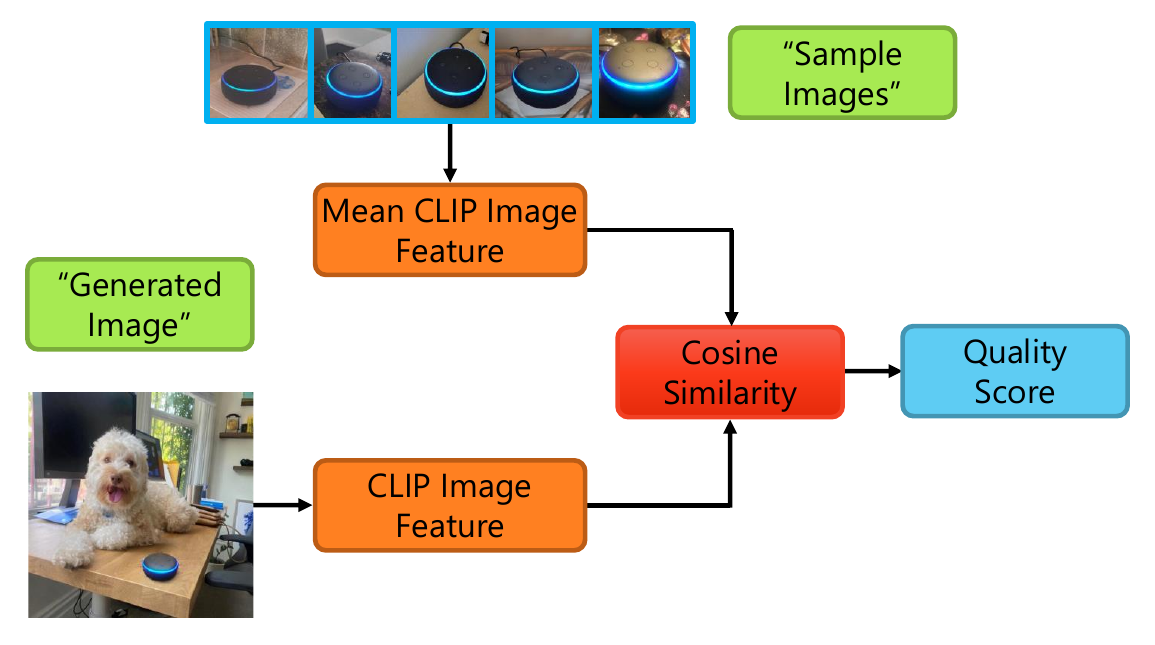}
    \caption{Quality Sub-module}
    \vspace{5pt}
\end{subfigure}

\begin{subfigure}[!htbp]{\columnwidth}
    \centering
    \includegraphics[width=0.7\columnwidth]{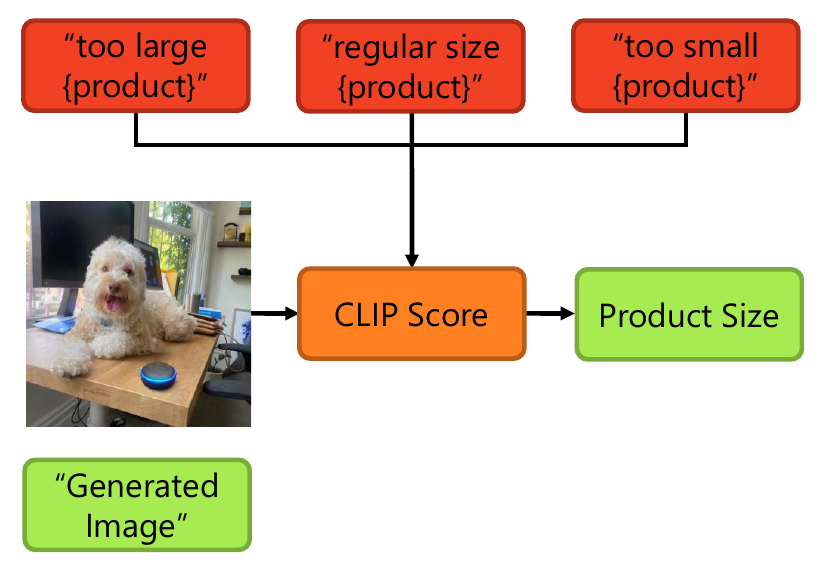}
    \caption{Volume Sub-module}
\end{subfigure}

\caption{Block diagram of each of the components of the Alignment Module. The Content sub-module is built using a pre-trained caption generator and CLIP models shown in (a). The generated caption is fine-tuned by adding the name of the intended product to the caption. For the Quality sub-module, the image features of the same CLIP model are utilized shown in (b). Finally, in the Volume sub-module, the same CLIP model with three different size text prompts is used shown in (c).}
\label{fig:alignment}
\end{figure}

\subsection{Product Alignment Module}
The Alignment Module comprises three sub-modules: Content, Quality, and Volume. The Content sub-module serves as a binary classifier, determining the presence of the product in the generated image. If the product's probability of existence surpasses a predefined threshold, then the Quality score is calculated for that image. This score evaluates the quality of the inpainted product in relation to the sample images originally used to train the SD model. Finally, if the image's quality score exceeds the set quality threshold, the Volume sub-module assesses the product's size in proportion to the background image. The generated image will be successfully accepted and presented to the user only if all three scores within the Product Quality Alignment Module meet their respective thresholds.
\par 
Within the Content module, an image captioning model \cite{imagecaption} is employed to generate a caption, which is then refined by incorporating the product's name. The super-class name of the product can also be utilized. Both the captions and the inpainted image are fed into the CLIP model to derive a CLIP score. If the modified caption scores above $70\%$, it's inferred that the product exists in the inpainted image.
The Quality module contrasts the mean CLIP image features of the sample images with the CLIP image feature of the generated image. The greater the resemblance of the inpainted product to the sample images, the higher the quality score. A threshold of $70\%$ has been established. 
The Volume module finally gauges the size of the inpainted product. The generated image is processed through the CLIP model, accompanied by three distinct textual size prompts. Given that size perception can be subjective and varies based on camera proximity, a milder threshold of $34\%$ (slightly above a random guess) has been selected. The comprehensive block diagram of the proposed VPP system is illustrated in \autoref{fig:block_diagram}, with the three stages distinguished by varied color blocks. The block diagrams for each sub-module can be found in \autoref{fig:alignment}.
\par
\begin{figure}[!htbp]
\centering
\includegraphics[width=\columnwidth]{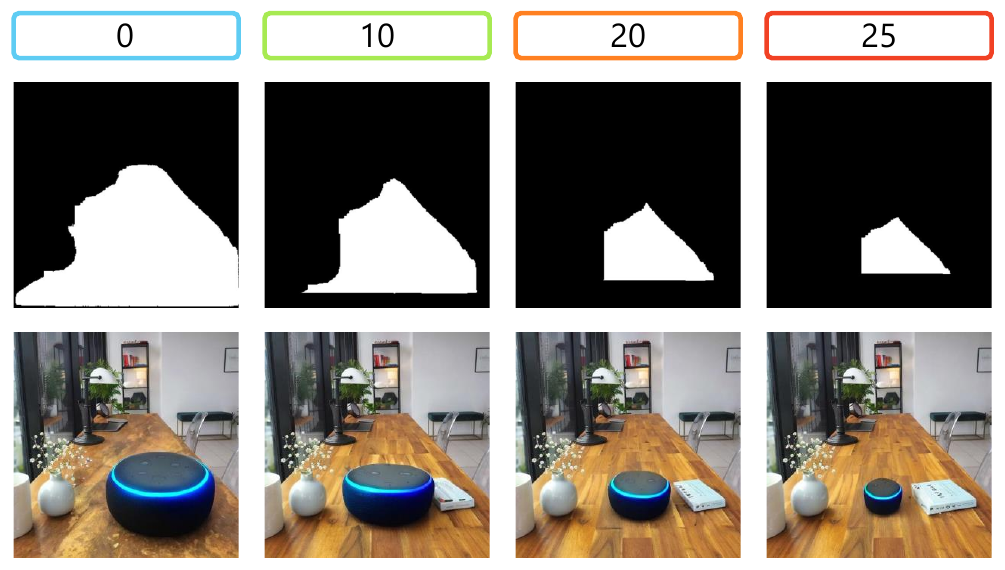}
\caption{Application of erosion to the mask where a kernel of size $(5\times5)$ is used for 0, 10, 20, and 25 iterations shown in the figure consecutively. The resulting output is presented at the bottom of the corresponding mask to show the size reduction of the generated product in the output image.}
\label{fig:erosion}
\end{figure}

 
The Volume sub-module provides insights regarding the size of the inpainted product. To modify the product's size, the mask's dimensions must be adjusted. For this task, morphological transformations, including mask erosion and dilation, can be employed on the binary mask. These transformations can either reduce or augment the mask area, allowing the inpainting module to produce a product image of the desired size. The relationship between alterations in the mask area and the size of the inpainted product across various erosion iterations is depicted in \autoref{fig:erosion}. Approximately, 25 iterations of erosion consume around 3 milliseconds, making it highly cost-effective.

\begin{table*}[!htbp]
\centering
\caption{Individual evaluation of content, quality, and volume sub-modules within the overall Alignment Module. ``Naive" represents the outputs without any filtering sub-modules. \emph{Content} classifies the presence of the product in the generated images. \emph{Quality} measures the proximity of the generated product to the sample product images used to fine-tune the diffusion model. Finally, \emph{Volume} identifies the size category of the product.}
\label{tab:scores}
\vspace{2mm}
\renewcommand{\arraystretch}{1.1}
\resizebox{\textwidth}{!}{%
\begin{tabular}{c|ccc|c|ccc|c|ccc|}
\cline{2-4} \cline{6-8} \cline{10-12}
 & & Naive & \textbf{Content} & \multirow{4}{*}{} &  & Naive & \textbf{Quality} & \multirow{4}{*}{} &  & Naive & \textbf{Volume} \\ \cline{2-4} \cline{6-8} \cline{10-12} 
% Amazon Echo Dot 
\multirow{3}{*}{\shortstack{Amazon\\Echo Dot}} & Success & 72 & 94 & \multirow{3}{*}{} & CLIP & $32.49 \pm 3.69$ & $33.80 \pm 2.69$ & \multirow{3}{*}{} & CLIP & $32.58 \pm 3.70$ & $33.42 \pm 2.69$ \\
 & Failure & 28 & 6 &   & MAQS & $4.41 \pm 3.23$ & $6.41 \pm 1.90$ &   & MASS & $3.01 \pm 2.68$ & $4.81 \pm 2.31$ \\
 & FR     & 38.89\% & 6.38\% &   &  MQS & $0.75 \pm 0.14$ & $0.83 \pm 0.06$   &    &  &  & \\ \cline{2-4} \cline{6-8} \cline{10-12}
% Lupure Vitamin C 
\multirow{3}{*}{\shortstack{Lupure\\Vitamin C}} & Success & 87 & 100 & \multirow{3}{*}{} & CLIP & $24.61 \pm 2.4$ & $25.23 \pm 2.66$ & \multirow{3}{*}{} & CLIP & $24.22 \pm 3.01$ & $24.51 \pm 2.89$ \\
 & Failure  & 13 &  0 &       & MAQS & $5.65 \pm 2.85$ & $6.47 \pm 1.09$ &  & MASS & $5.64 \pm 3.05$ & $7.14 \pm 1.53$ \\
 & FR  & 14.94\% & 0.0\% &   &  MQS & $0.81 \pm 0.13$ & $0.86 \pm 0.04$ &  &  &  & \\ \cline{2-4} \cline{6-8} \cline{10-12} 
\end{tabular}%
}
\end{table*}

\section{Experimental Results} \label{sec:experiment}
Experiments were conducted to evaluate the performance of the proposed VPP system. For these experiments, five sample images of an ``Amazon Echo Dot" were chosen. $1,000$ augmented images of each product created from these five sample images were used to fine-tune the DreamBooth model using the text prompt "A photorealistic image of a sks Amazon Alexa device." The model was fine-tuned for $1,600$ steps, employing a learning rate of $5 \times 10^{-6}$, and a batch size of 1.
\par 
The fine-tuned model can inpaint products into the masked region. However, issues such as lack of product appearance, poor resolution, and disproportionate shape persist. The goal of the proposed Alignment Module is to automatically detect these issues. If identified, the problematic images are discarded, and a new image is generated from different random noise. Only if a generated image meets all the module's criteria it is presented to the user. Otherwise, a new image generation process is initiated. This loop continues for a maximum of 10 iterations.

\subsection{Assessing Alignment Module}

To assess the effectiveness of the Alignment Module, images were generated both with and without it. For each sub-module, as well as for the overall Alignment Module, 200 images were generated: 100 with the filter activated and 100 without (referred to as the "Naive" case). 
\par 
To prevent bias, all images were given random names and were consolidated into a single folder. These images were also independently evaluated by a human, whose scores served as the ground truth. This ground truth information was saved in a separate file for the final evaluation, which followed a blindfolded scoring method. All the experiments were also repeated for another product named ``Lupure Vitamin C".

\begin{table*}[!htbp]
\centering
\caption{Comparison of the proposed method with and without using the Alignment Module in addition to the Paint-By-Example (PBE) \cite{yang2023paint} inpainting model. The ``Naive" performance represents the generated output without applying the Alignment Module. The ``Alignment" column represents the generated outputs where three cascaded filtering sub-modules are used, i.e., the Alignment Module.}
\label{tab:alignment}
\vspace{2mm}
\renewcommand{\arraystretch}{1.0}
\resizebox{0.8\textwidth}{!}{%
\begin{tabular}{@{}ccccccc@{}}
\toprule
 & \multicolumn{3}{c}{\shortstack{Amazon Echo Dot}} & \multicolumn{3}{c}{\shortstack{Lupure Vitamin C}} \\ \midrule
 & PBE & Naive & \textbf{Alignment} & PBE & Naive & \textbf{Alignment} \\ \midrule
CLIP & $31.44 \pm 3.43$ & $32.85 \pm 3.19$ & $33.85 \pm 2.54$ & $27.01 \pm 2.10$ & $24.71 \pm 2.64$ & $24.89 \pm 2.90$ \\
MAQS & $1.13 \pm 1.30$ & $4.65 \pm 3.60$ & $6.31 \pm 2.39$ & $1.75 \pm 1.51$ & $6.60 \pm 3.01$ & $7.81 \pm 1.13$ \\
MASS & $1.22 \pm 1.60$ & $3.05 \pm 2.98$ & $4.70 \pm 2.81$ & $2.43 \pm 2.07$ & $6.25 \pm 3.08$ & $7.30 \pm 1.59$ \\ 
MQS  & $0.64 \pm 0.08$ & $0.75 \pm 0.14$ & $0.82 \pm 0.05$ & $0.67 \pm 0.06$ & $0.82 \pm 0.12$ & $0.86 \pm 0.05$ \\ 
FR   & $78.57\%$ & $29.87\%$ & $0.00\%$ & $38.89\%$ & $17.64\%$ & $0.00\%$ \\ \bottomrule
\end{tabular}%
}
\end{table*}

\subsection{Evaluation Metrics} 
The evaluation and scoring method of each of the sub-modules of the Alignment module is described in the consecutive segments.
\vspace{5pt}

\begin{itemize}
  \item \textbf{Content Score} \quad For the image content score, images are categorized into two classes: `success' if the product appears, and `failure' otherwise. When the content module is utilized, the Failure Rate (FR), defined as the ratio of Failure to Success, is below $10\%$ for both of the products. 
  
  \vspace{5pt}
  
  \item \textbf{Quality Score} \quad For the quality score, images are rated on a scale from 0 to 10: 0 indicates the absence of a product, and 10 signifies a perfect-looking product. To evaluate in conjunction with the CLIP score, both the Mean Assigned Quality Score (MAQS) and Mean Quality Score (MQS) are calculated. MAQS represents the average score of images labeled between 0 and 10, while MQS is the output from the quality module, essentially reflecting cosine similarity.
  
  \vspace{5pt}
  
  \item \textbf{Volume Score} \quad For the volume module, images are also rated on a scale from 0 to 10: 0 for a highly unrealistic size, and 10 for a perfect size representation. When evaluating the volume module, the content module is not utilized. Since the size score necessitates the presence of a product, images without any product are excluded from this evaluation. To gauge performance, the Mean Assigned Size Score (MASS) is calculated in addition to the CLIP score.
\end{itemize}

\begin{figure}[!t]
\centering
\begin{subfigure}[!htbp]{0.33\columnwidth}
    \centering
    \includegraphics[width=\columnwidth]{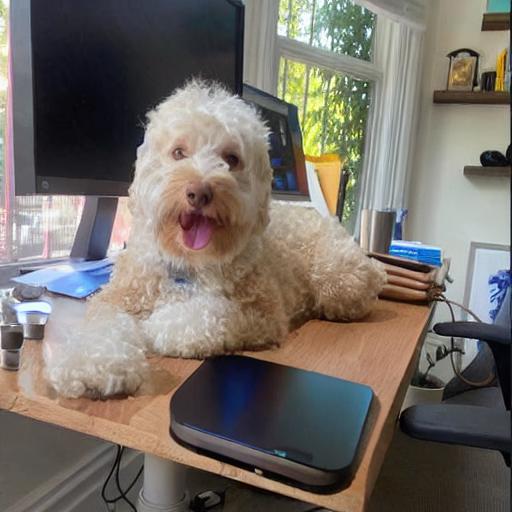}
    \caption{}
\end{subfigure}
\hspace{1.9mm}
\begin{subfigure}[!htbp]{0.33\columnwidth}
    \centering
    \includegraphics[width=\columnwidth]{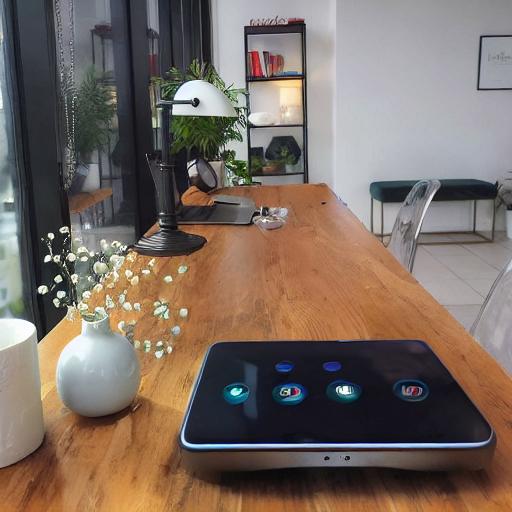}
    \caption{}
\end{subfigure}

\begin{subfigure}[!htbp]{0.33\columnwidth}
    \centering
    \includegraphics[width=\columnwidth]{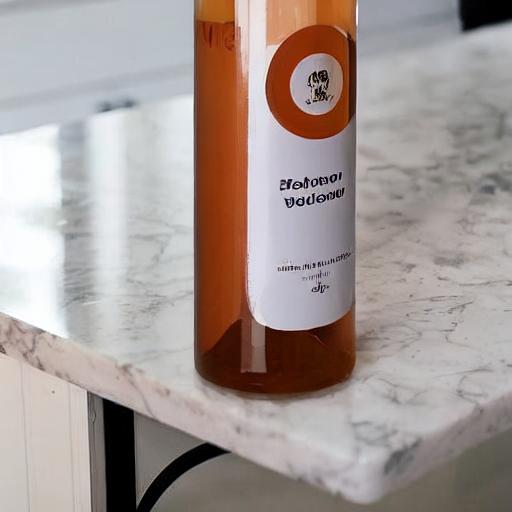}
    \caption{}
\end{subfigure}
\hspace{1.9mm}
\begin{subfigure}[!htbp]{0.33\columnwidth}
    \centering
    \includegraphics[width=\columnwidth]{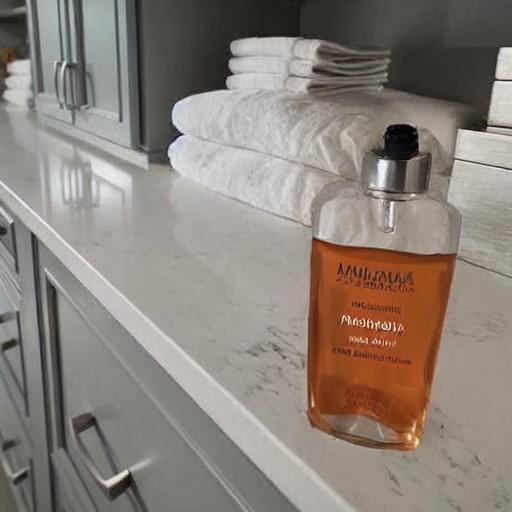}
    \caption{}
\end{subfigure}
\caption{Inpainted product image of Paint-by-Example (PBE). PBE generates high-quality images which explains the higher CLIP score in the case of Lupure Vitamin C. However, the inpainted product does not look similar to the desired product at all resulting in very poor mean assigned quality and size scores. Output images for Amazon Echo Dot is shown in (a) and (b), and for Lupure Vitamin C is shown in (c) and (d).}
\label{fig:pbe}
\end{figure}

\begin{figure}[!htbp]
\centering
\includegraphics[width=\columnwidth]{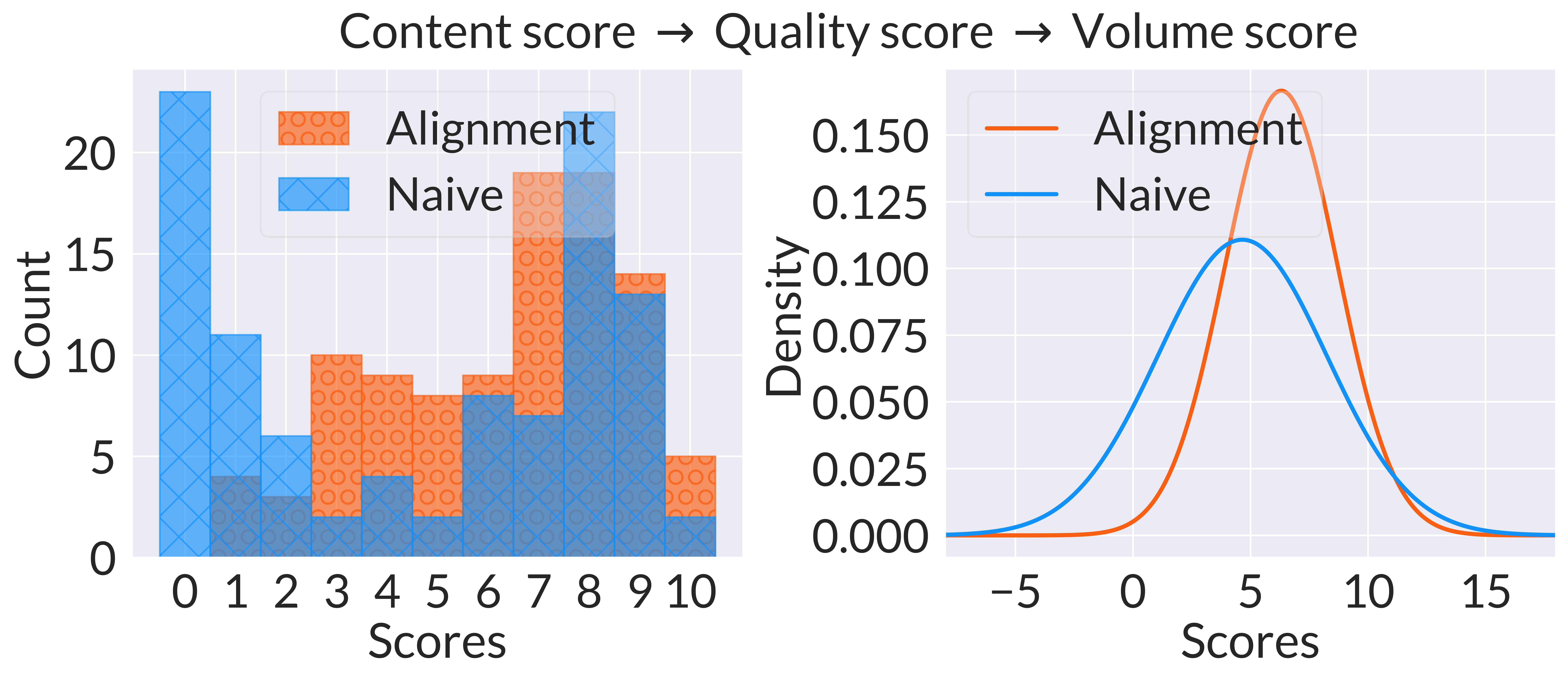}
\caption{Empirical performance of Alignment Module for Amazon Echo Dot. Noticeably, no output is generated without any product when the Alignment Module is employed. Moreover, the mean quality score has increased from 4.65 to 6.31.}
\label{fig:alignment_eval}
\end{figure}

\subsubsection{Overall Results}
The results of individual evaluations are presented in \autoref{tab:scores}. It can be observed from this table that using any of the sub-modules consistently produced better outcomes compared to when no filtering was applied across various metrics. The results of the comprehensive evaluation, encompassing all sub-modules, can be found in \autoref{tab:alignment}.

\subsection{Comparison with Paint-By-Example}
\begin{figure*}[!t]
\centering
\begin{subfigure}{\textwidth}
    \includegraphics[width=\textwidth]{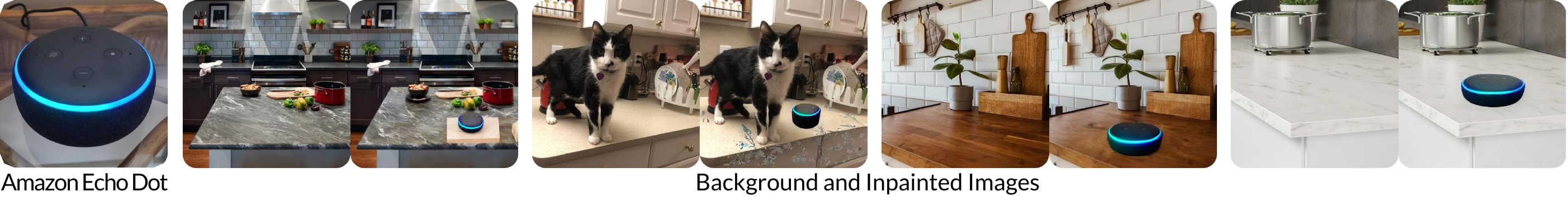}
    \vspace{1pt}
\end{subfigure}
\begin{subfigure}{\textwidth}
    \includegraphics[width=\textwidth]{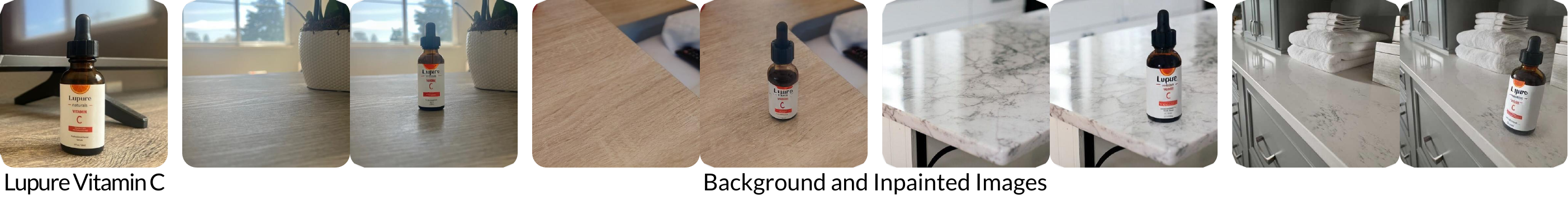}
\end{subfigure}
\caption{Qualitative results of the proposed VPP system. Experiments are performed using two different products, Amazon Echo Dot as shown on top, and Lupure Vitamin C as shown on bottom. The original training images are shown on the left, and then the pairs of background and inpainted output images are presented side by side.}
\label{fig:outputs}
\end{figure*}

The proposed method is compared with the Paint-By-Example (PBE) \cite{yang2023paint} inpainting model and \autoref{tab:alignment} shows the performance comparison of the proposed method along with PBE. PBE can generate very high-quality images, however, the inpainted product in the generated image does not look alike the desired product at all as shown in \autoref{fig:pbe} resulting in very poor MAQS and MASS. Whereas the inpainted product of our proposed method resembles much of the original product shown in Figure \autoref{fig:outputs}. 

\subsection{Frequency Distribution}
The frequency distribution and density function of the assigned quality scores in the case of ``Naive" and ``Alignment" for Amazon Echo Dot is presented in \autoref{fig:alignment_eval}. The density mean has shifted from 4.65 to 6.31 when Alignment Module is adopted indicating the effectiveness of the proposed module.

\begin{figure*}[!htbp]
\centering
\includegraphics[width=0.6\textwidth]{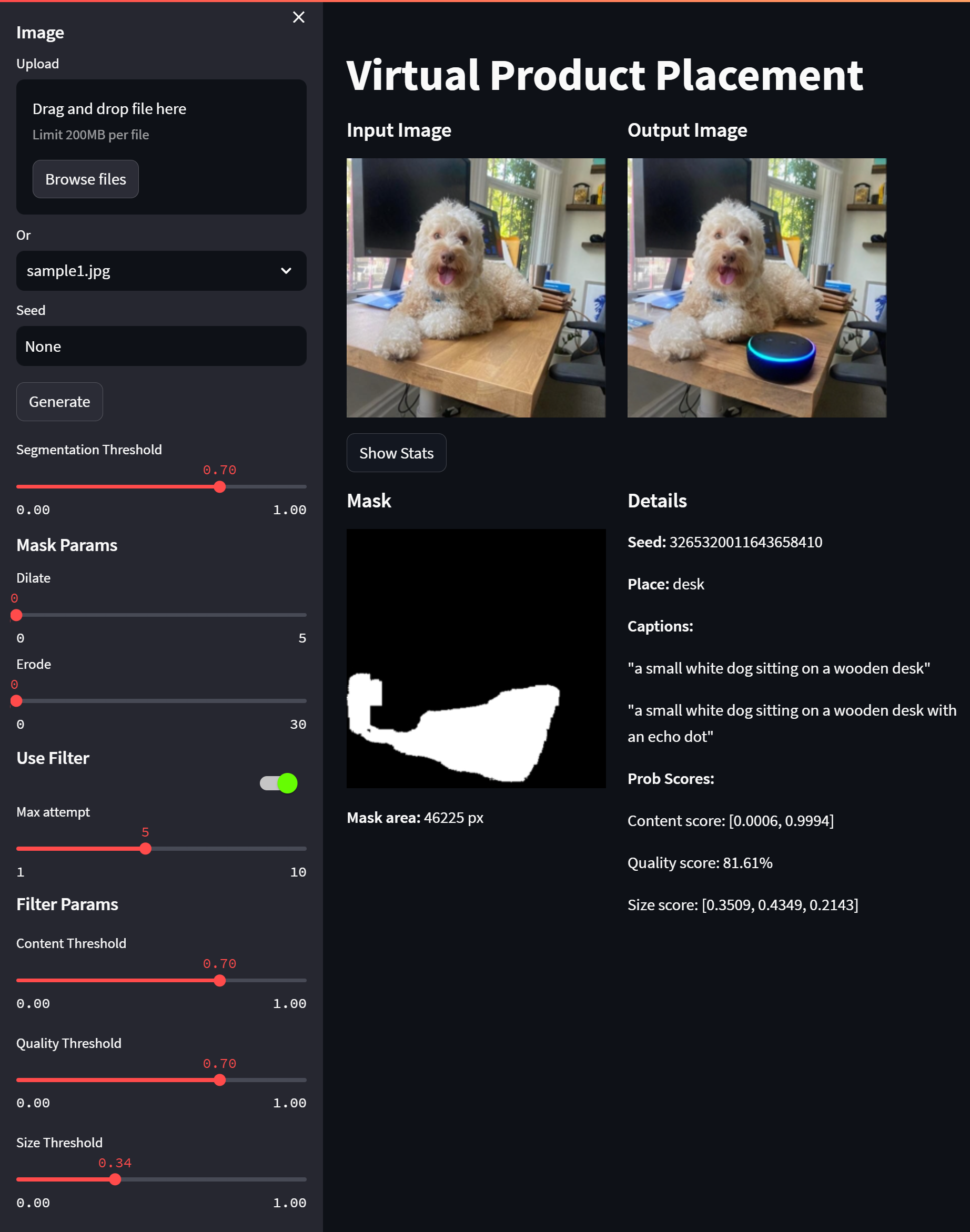}
\caption{The interface of the VPP web app demo was built using Streamlit hosted in Amazon SageMaker. The uploaded background image is shown under the title ``Input Image" and the inpainting image with an Amazon Echo Dot is shown under the title ``Output Image". Moreover, the generated mask produced by the location identifier and the other intermediate details of the proposed VPP system is also presented in the interface.}
\label{fig:app}
\end{figure*}

\section{Path to Production} \label{sec:production}
% 1) explain your API, put a snapshot of the demo
% 2) talk about all the practical considerations that should be kept into account
% 3) talk about testing in production and potentially the possibility of a human in the loop
\subsection{Product API}
The location identifier, fine-tuned model, and Alignment Module are combined to develop an easy-to-use VPP Streamlit web app\,\footnote{\textsc{Streamlit}: \texttt{https://streamlit.io/}}. This app is hosted on Amazon Sagemaker using an ``ml.p3.2xlarge" instance, which is a single V100 GPU with 16GB of GPU memory. The demo app's interface is illustrated in \autoref{fig:app}. In the top-left `Image' section, users can either upload their own background image or choose from a selection of sample background images to generate an inpainted product image. 
\par 
The web app provides extensive flexibility for tuning the parameters of the Alignment Module so that users can comprehend the effects of these parameters. In the `seed' text box, a value can be input to control the system output. The segmentation threshold for CLIPSeg defaults to 0.7, but users can refine this value using a slider. Within the `Mask Params' section, the number of dilation and erosion iterations can be set and visualized in real-time. 
\par 
The filter, represented by the Alignment Module, can be toggled on or off. The `Max Attempt' slider determines the number of regeneration attempts if the model doesn't produce a satisfactory output. However, if a seed value is specified, the model will generate the output only once, regardless of the set value. Lastly, in the `Filter Params' section, users can fine-tune the threshold values for each sub-module of the Alignment Module, specifically for content, quality, and volume. 
\par 
The ``show stats" button beneath the input image displays the mask alongside details of the model outputs. These details include the seed value, placement, generated and modified captions, and the content, quality, and volume/size scores. By visualizing the mask and its area, users can apply erosion or dilation to adjust the product's size. The default threshold values for content, quality, and volume are $0.7$, $0.7$, and $0.34$, respectively. While these values can be adjusted slightly higher, it's recommended to also set the 'Max Attempt' to 10 in such cases. A higher threshold means that the generated output is more likely to fail the criteria set by the Alignment Module.

\subsection{Future Considerations for Product Scalability}
Fine-tuning stable diffusion using DreamBooth can take up to 30 minutes, depending on dataset size, image resolution, and extent of training. When considering a customer with hundreds or thousands of products, this process could take days to complete model training across different products. 
\par 
Our pipeline is deployed on Amazon SageMaker, a managed service that supports the automatic scaling of deployed endpoints. This service can dynamically accommodate large computational needs by provisioning additional instances as required. As such, fine-tuning 100 SD models for 100 different products would still only take about 30 minutes if 100 instances were utilized in parallel. 
\par 
The fine-tuned models are stored in an Amazon S3 (Simple Storage Service) bucket, with each model being ~2.2 GB in size. Consequently, 100 fine-tuned models would occupy approximately 220 GB of storage space. A pertinent question arises: Can we strike a space-time trade-off by training a single model with a unique identifier for each product? 
\par 
If this is feasible, the space requirement would be reduced to a consistent 2.2 GB. However, that one model would need more extensive training - specifically training steps would increase by a factor of 100 for 100 products, thereby lengthening the computation time. This approach remains untested and warrants future exploration \cite{kumari2023multi}.

\section{Conclusion} \label{sec:conclusion}
In this paper, we present a novel, fully automated, end-to-end pipeline for Virtual Product Placement. The proposed method automatically determines a suitable location for product placement into a background image, performs product inpainting, and finally evaluates image quality to ensure only high-quality images are presented for the downstream task.
\par 
Using two different example products, experiments were conducted to evaluate the effectiveness of the proposed pipeline, the performance of the individual sub-modules, and the overarching Alignment Module. Notably, when upon employing the Alignment Module, the Failure Ratio (FR) plummeted down to $0.0\%$ for both investigated products. Additionally, images produced with the Alignment Module achieved superior CLIP, quality, and size scores. 
\par 
Qualitatively, the produced images present a clean and natural semantic inpainting of the product within the background image. The accompanying web application facilitates pipeline deployment by enabling image generation through a user-friendly interface with extensive image fine-tuning capabilities. The high-quality integration of products into images underscores the potential of the proposed VPP in the realms of digital marketing and advertising.

\small
\bibliographystyle{ieeenat_fullname}

\end{document}

%% file: preamble.tex
\usepackage[dvipsnames]{xcolor}